\documentclass[runningheads]{llncs}
\usepackage[T1]{fontenc}
\usepackage{graphicx}

\usepackage{hyperref}
\usepackage{color}

\usepackage{booktabs}
\usepackage{amssymb}
\usepackage{amsmath}
\usepackage{cleveref}
\begin{document}
\title{Segmentation-Assisted Brain MRI Synthesis with Cross-Image Multi-Contrast Feature Memory
Bank Retrieval Augmentation}
\titlerunning{Segmentation-Assisted Brain MRI Synthesis}
%
\author{Wenwei Huang\inst{1}\orcidID{0009-0008-9907-6801} \and
Jia Wei\inst{2}\orcidID{0000-0002-5794-1712}\thanks{Corresponding author}\and
Jianlong Zhou\inst{3}\orcidID{0000-0001-6034-644X}}
\authorrunning{W. Huang et al.}
%
\institute{South China University of Technology, Guangzhou 510640, China \email{1942569178@qq.com} \and
South China University of Technology, Guangzhou 510640, China
\email{csjwei@scut.edu.cn}\\
\and
UTS Data Science Institute, University of Technology Sydney, Ultimo, NSW 2007, Australia\\
\email{jianlong.zhou@uts.edu.au}}
\maketitle              
\begin{abstract}
Multi-contrast brain MRI provide complementary soft-tissue characteristics that aid in the screening and diagnosis of diseases. However, limited scanning time, image corruption and various imaging protocols often result in incomplete multi-contrast images. While current approaches excel in image synthesis, they often struggle to synthesize critical tumor regions and exploit contextual information in multi-contrast brain MRI effectively. To address this issue, we propose a synthesis-centric, segmentation-assisted closed-loop framework with retrieval augmentation synthesis. Our method overall takes a generative adversarial architecture, which aims to synthesize missing contrasts from any combination of available ones with a single model. To explicitly capture tumor semantics and focus synthesis on tumor regions, we add an auxiliary segmentation branch that predicts tumor masks and feeds them back as semantic conditioning in synthesis branch, thereby learning tumor-aware representations in the model and improving synthesis fidelity. Furthermore, we propose a dual-bank retrieval augmentation strategy. It dynamically queries two external knowledge bases, namely a tumor masks memory bank for crucial tumor context and cross-image contrast feature memory bank for global style information, to augment synthesis. Verified on two public multi-contrast magnetic resonance brain datasets: BraTs2020 and UCSF-BMSR, the proposed method is effective in handling medical brain images synthesis tasks and shows superior performance compared to previous methods. Code is available at: \url{https://github.com/iBizzard/SSCF.git}

\end{abstract}
\section{Introduction}

Multi-contrast brain MRI provides complementary diagnostic information for disease assessment. In routine clinical practice, T1-weighted (T1w), T2-weighted (T2w), contrast-enhanced T1-weighted (T1ce), and fluid-attenuated inversion recovery (FLAIR) images characterize different anatomical structures and pathological patterns. For brain gliomas, radiological evaluation mainly focuses on three critical regions: peritumoral edema (ED), enhancing tumor (ET), and non-enhancing tumor core (NET). In particular, T2w is sensitive to edema, while T1ce better highlights enhancing lesions and necrotic components \cite{Alpher33}. However, due to limited scan time, patient motion, and imaging artifacts, acquiring a complete set of contrasts is often difficult in practice \cite{Alpher02}. This limitation has motivated substantial interest in multi-contrast image synthesis, which aims to infer missing contrasts from available ones to support downstream tasks such as segmentation \cite{Alpher03}, registration \cite{Alpher04}, diagnosis, and treatment planning \cite{Alpher06}.

Recent multi-contrast synthesis methods are mainly based on generative models, including GANs \cite{Alpher07} and diffusion models \cite{Alpher35}. Early GAN-based approaches explored contrast synthesis through disentangled representation learning, unified generators, adaptive fusion, and graph-based hyper-encoders \cite{Alpher02,Alpher09,Alpher13,Alpher11}. Transformer-based methods further improved the modeling of long-range inter-contrast dependencies and multi-scale feature alignment \cite{Alpher17,Alpher18,Alpher19}. More recently, diffusion models have shown strong perceptual quality in medical image synthesis \cite{Alpher29}. Despite this progress, most existing methods primarily exploit inter-contrast correlations. When multiple contrasts are missing, such dependence often becomes unreliable, leading to blurred details or anatomically inconsistent structures, especially in clinically important tumor regions.

Another limitation of current methods is that they rarely leverage cross-image contextual knowledge from other subjects. In fact, external samples may provide useful style and texture priors when the available contrasts in the current subject are insufficient. This issue is related to the broader success of memory-based non-parametric learning in computer vision, where external feature banks have been used to enhance restoration, salient object detection, few-shot generation, and image translation \cite{Alpher20,Alpher21,Alpher22,Alpher23,Alpher24,Alpher25,Alpher27}. In medical imaging, memory-bank paradigms are also emerging as a promising strategy \cite{Alpher08}. However, existing memory-based approaches usually treat images independently or focus on single-modality settings, and therefore do not fully exploit the structural complementarity inherent in multi-contrast MRI.

In addition, most synthesis frameworks impose insufficient explicit constraints on tumor structure. Since tumor morphology strongly constrains the appearance of different MRI contrasts, tumor semantics can serve as effective structural priors for synthesis. Motivated by this observation, we propose a synthesis-centric, segmentation-assisted closed-loop framework with retrieval-augmented synthesis. Specifically, a segmentation branch predicts tumor masks and feeds them back to the synthesis branch as semantic guidance, improving structural fidelity and fine-grained tumor texture reconstruction. To compensate for missing information, we further construct a cross-image multi-contrast feature memory bank to provide external style and texture priors, as well as a tumor masks memory bank to model cross-image tumor relationships and support more reliable segmentation.

The main contributions are as follows:

1. We propose a synthesis-centric, segmentation-assisted closed-loop framework (SSCF), in which the segmentation branch provides tumor guidance to the synthesis branch, explicitly improving structural consistency and fine-grained tumor region reconstruction.

2. We introduce a retrieval-augmented synthesis (RAS) strategy with a cross-image multi-contrast feature memory bank, enabling the model to retrieve relevant external priors and enhance synthesis fidelity under severe contrast missingness.

3. We design a tumor masks memory bank for cross-image tumor information modeling, which works jointly with the feature memory bank to strengthen tumor-aware representation learning and improve masks prediction accuracy.

\begin{figure*}
  \centering
  \includegraphics[width=\linewidth]{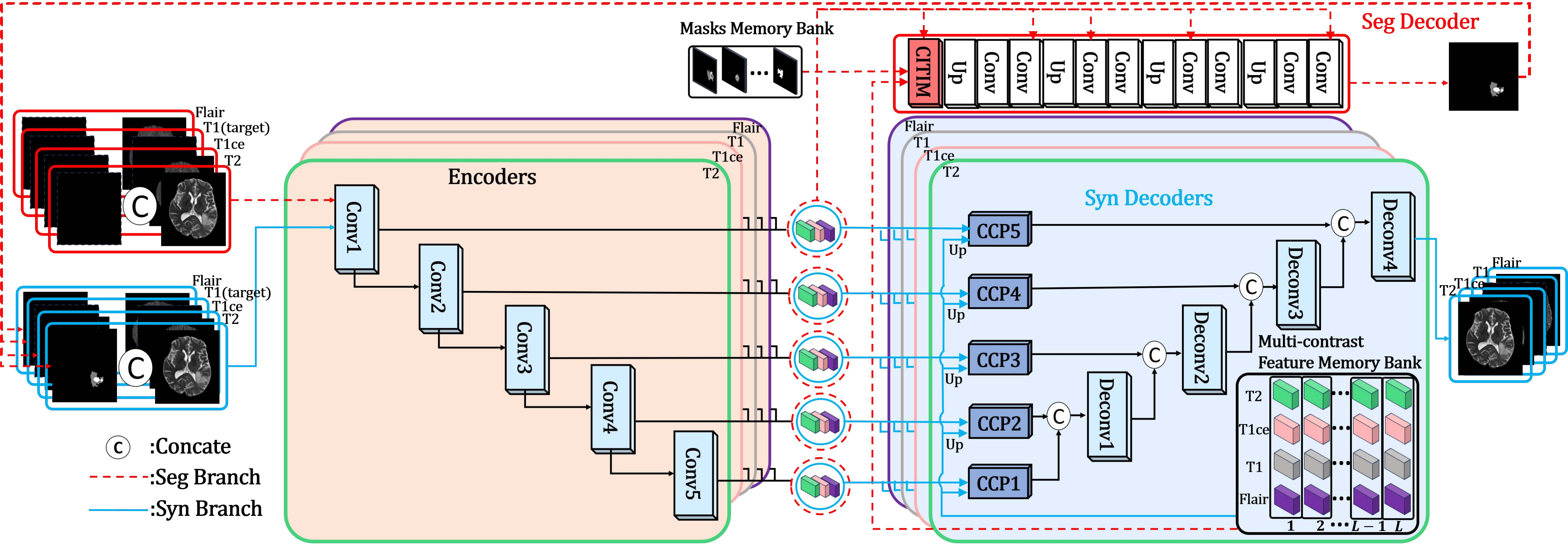}
  \hfill
  \caption{Overview of the proposed framework. Our model consists of segmentation and synthesis branch. (a) Segmentation Branch (red): This branch encodes available contrasts and leverages a Cross-Image Tumor Information Modeling (CITIM) module to produce accurate tumor masks via the seg decoder. (b) Synthesis Branch (blue): Taking the available contrasts and predicted masks as input, the branch employs Cross-Contrast Prediction (CCP) modules to drive the synthesis of the target contrasts and the reconstruction of available contrasts through four parallel syn decoders. }
  \label{Figure:a}
\end{figure*}

\section{Methodology}
Our framework performs any-to-any multi-contrast brain MRI synthesis using complete training subjects with tumor masks. During training, the full contrast set $X=\{x^i \mid i \in \{Flair,T1,T1ce,T2\}\}$ is randomly split into available contrasts $\mathcal{S}$ and target contrasts $\mathcal{T}$, where $\mathcal{S}\cup\mathcal{T}=X$ and $\mathcal{S}\cap\mathcal{T}=\emptyset$. As shown in \Cref{Figure:a}, the model contains a segmentation branch and a synthesis branch sharing encoders $E^i$. The segmentation branch predicts tumor masks $\tilde{y}$, which are then injected into the synthesis branch as semantic guidance for target contrast generation.

During training, available and target contrasts are encoded, but only features from $\mathcal{S}$ are fed to the segmentation and synthesis decoders. Features from $\mathcal{T}$ are used only as ground-truth representations for feature alignment, forcing the model to infer target features from available inputs. The whole framework is optimized end-to-end, allowing synthesis losses to improve the shared tumor-aware representations.

To capture contrast-specific style statistics and tumor semantics, the model maintains two FIFO-updated cross-image memory banks: a multi-contrast feature memory bank $B_f \in \mathbb{R}^{L \times 4 \times B \times C \times h \times w}$ and a tumor masks memory bank $B_m \in \mathbb{R}^{L \times c \times h \times w}$, where $c$ is the number of tumor classes. We use the Cross-Image Tumor Information Modeling (CITIM) module in $D_{\text{seg}}$ and the Cross-Contrast Prediction (CCP) module in $D_{\text{syn}}^i$ to retrieve tumor-aware and contrast-aware priors. Four PatchGAN discriminators \cite{Alpher28} are used to enforce high-frequency realism.

\begin{figure*} 
  \centering
  \includegraphics[width=\linewidth]{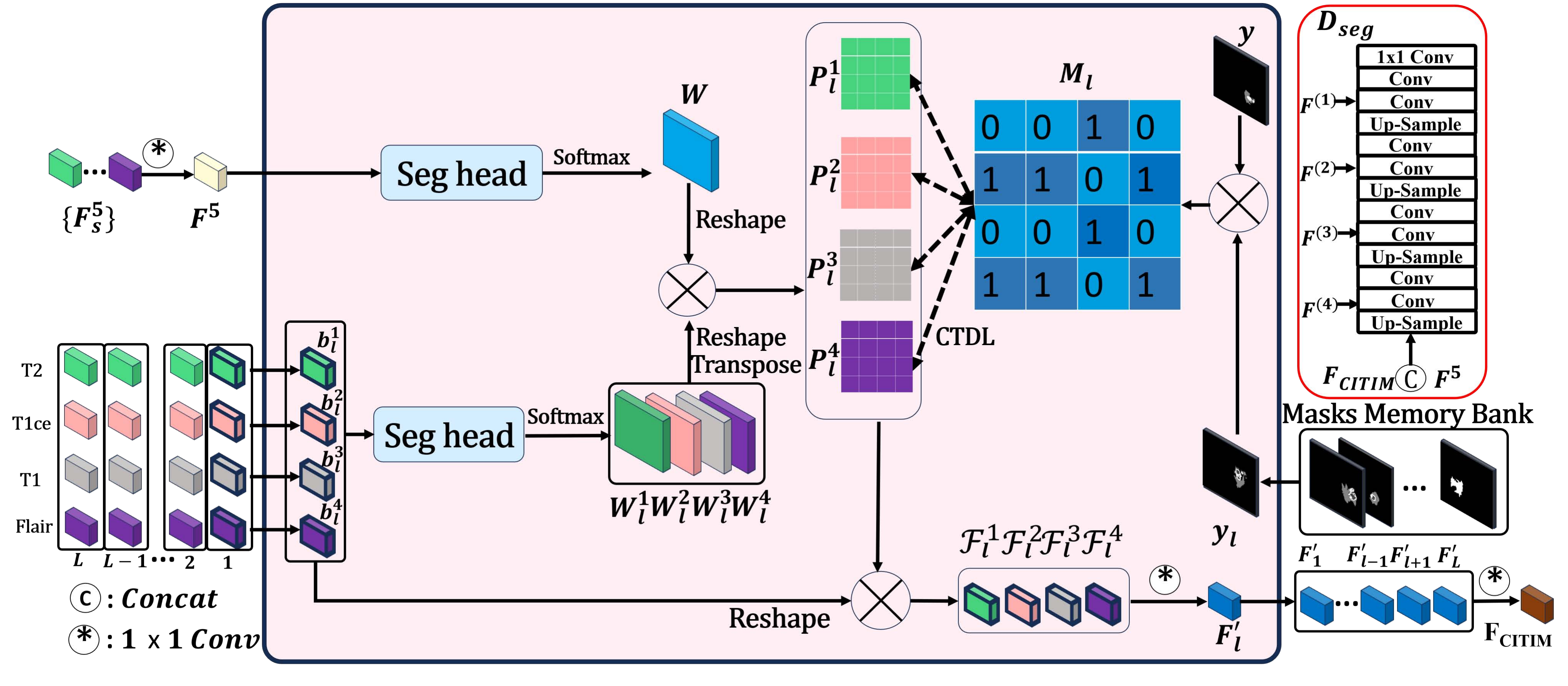} 
  \caption{Overview of the CITIM. This module models cross-image dependencies and extracts tumor-specific features by leveraging bottleneck features from the encoders, cross-image multi-contrast feature memory bank, tumor masks and tumor masks memory bank as input. }
  \label{fig:2}
\end{figure*}

\subsection{Segmentation Branch in SSCF}

As depicted in \Cref{fig:2}, in the segmentation branch, the encoders $E^{s}$ extract multi-scale features, denoted as $\{F_{s}^j\}$ where $j \in \{1, \dots, 5\}$ represents the scale and $s \in \mathcal{S}$. At the bottleneck of $D_{\text{seg}}$, the deepest features $\{F_{s}^5\}$ are fused via a $1\times 1$ convolution to produce a unified representation $F^{5}$. This fused feature, along with the tumor masks memory bank and cross-image multi-contrast feature memory bank, serves as input to the CITIM module.

The CITIM module is designed to capture cross-image local tumor context. Acknowledging the distinct contribution of each contrast to tumor definition, we model their effects independently before aggregation. Let $b_l$ denote the $l$-th feature quadruplet, which consists of four contrast features from a single subject and is stored in $B_f$. First, a lightweight prediction head (four convolution blocks with Sigmoid activation) projects both $F^{5}$ and $b_l$ into class-probability maps $W \in \mathbb{R}^{{c} \times h \times w}$ and $W_{l} \in \mathbb{R}^{4 \times{c} \times h \times w}$ respectively. 

To define semantic correspondence, we flatten the spatial dimensions to $N = h \times w$. A cross-image affinity matrix $P_{l} \in \mathbb{R}^{4\times{N} \times N }$ is computed to measure pixel-level semantic similarities between the current input and the memory subject:
    \begin{equation}
      P_l = W^T \otimes W_l 
      \label{eq:pl_matrix}
    \end{equation}
where ${\otimes}$ is matrix multiplication, and for a given contrast, the element $p_{mn}$ of $P_l$ represents the semantic similarity between location $m$ in the memory subject and location $n$ in the current subject. This matrix is then used to retrieve and aggregate relevant features from the memory feature $b_l$ after flatten, producing a cross-image enhanced feature $\mathcal{F}_l$:
\begin{equation}
  \mathcal{F}_l = \text{Norm}(b_l \otimes P_l) \in \mathbb{R}^{4 \times B\times{C} \times h \times w}
  \label{eq:feat_agg}
\end{equation}
where $\text{Norm}(\cdot)$ denotes Layer Normalization. To integrate contributions from all contrasts, the retrieved features are concatenated and fused:
\begin{equation}
  \mathcal{F}_l^{'} = \psi \big(\mathcal{F}_l \big) \in \mathbb{R}^{ B\times{C} \times h \times w}
  \label{eq:fusion_l}
\end{equation}
This process iterates over all $L$ subjects in the memory bank, and the results are further aggregated to form the final CITIM representation:
\begin{equation}
  F_{\text{CITIM}} = \phi \big(\text{Concat}(\mathcal{F}_1^{'}, \mathcal{F}_2^{'}, \dots, \mathcal{F}_L^{'}) \big) \in \mathbb{R}^{ B\times{C} \times h \times w}
  \label{eq:fcmm}
\end{equation}
where $\psi(\cdot)$ and $\phi$ denote a $1\times 1$ convolution. To supervise this dependency modeling, we construct an ideal dependency map $M_l \in \mathbb{R}^{N \times N}$ using the ground-truth masks $y$ and the memory masks $y_l$:
\begin{equation}
  M_l = y^T \otimes y_l
  \label{eq:ml}
\end{equation}
the Cross-Image Tumor Dependency Loss (CTDL) \cite{Alpher10} is then applied to regularize $P_l$:
\begin{equation}
  \ell{_{c}^l} = -\frac {1}{4 N^{2}} \sum_{i=1}^{4}\sum_{k=1}^{N^{2}} (m_k \log p_k + (1 - m_k) \log (1 - p_k))
  \label{eq:lc_single}
\end{equation}
The total CTDL loss is averaged over the bank: 
\begin{equation}
\ell{_{c}} = \frac{1}{L} \sum_{l=1}^{L} \ell{_c^l}
\end{equation}
This mechanism ensures that $F_{\text{CITIM}}$ captures enriched tumor semantics, providing discriminative features essential for segmentation, even in the presence of missing contrasts. Finally, $F_{\text{CITIM}}$ is combined with $F^{5}$ and fed into $D_{\text{seg}}$ to predict $\tilde{y}$.

\subsection{Synthesis Branch in SSCF}
 
\begin{figure} 
  \centering
  \includegraphics[width=0.6\linewidth]{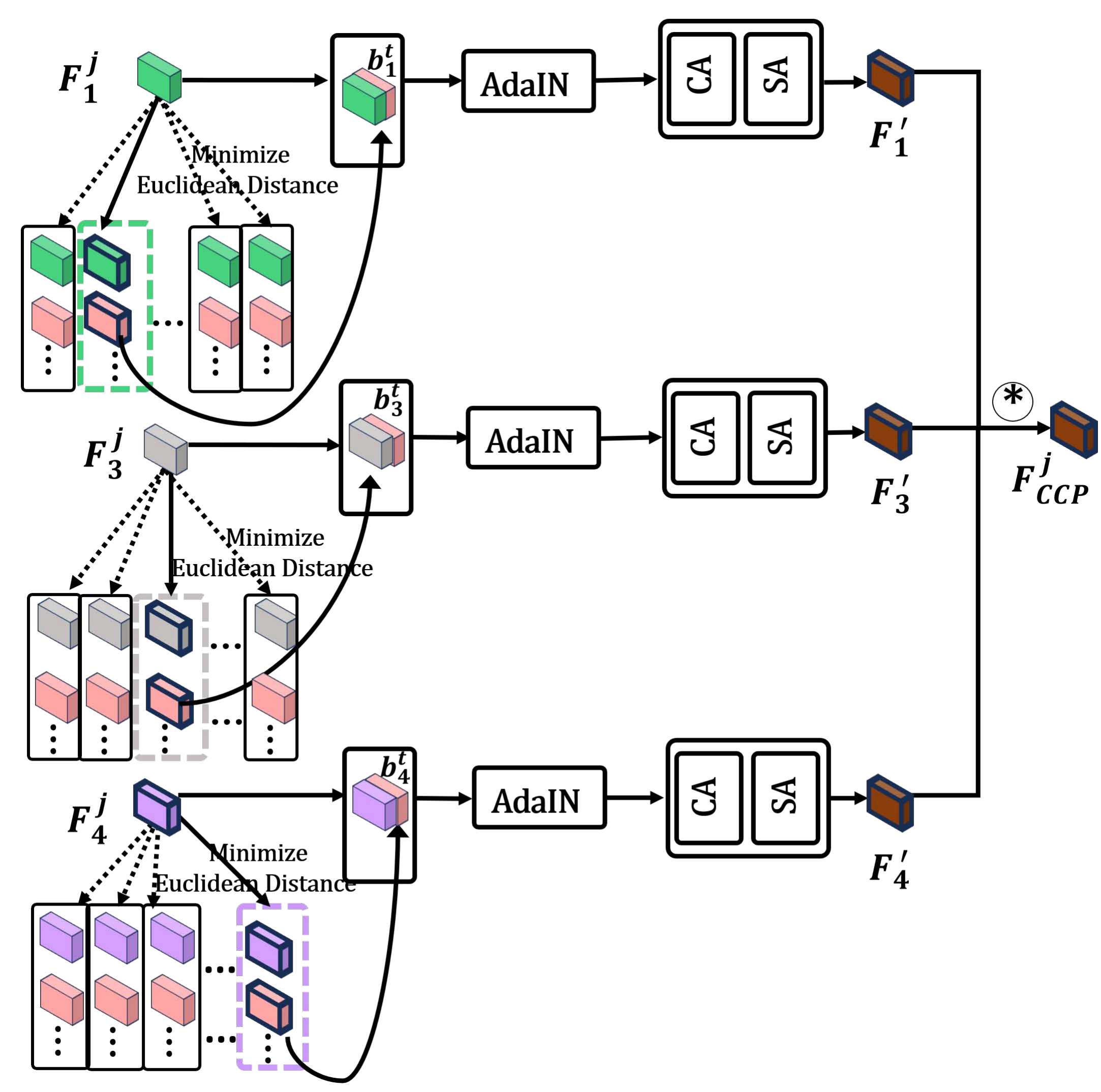} 
  \caption{Overview of the CCP. The module predicts target contrast feature by retrieving information from the corresponding cross-image multi-contrast feature memory bank. This process is guided by features from available contrasts $\{F_s^j\}$, and the retrieved information is enhanced alongside them before the final prediction. }
  \label{fig:3}
\end{figure}

The synthesis branch shares the encoders with the segmentation branch and takes available contrasts $X^s$ together with the predicted masks $\tilde{y}$ as input. To reduce contrast distribution gaps, each target contrast is generated by an independent decoder.

\textbf{Retrieval Augmented Synthesis}. As illustrated in \Cref{fig:3}, CCP modules are inserted at five scales to retrieve cross-image priors for missing contrasts. At scale $j$, the available feature $F_s^j$ is used as the query, and Euclidean distance is used to find the most similar subject index $l^*$ from the feature memory bank. The corresponding target contrast feature $F_{l^*}$ is retrieved as a prototype that provides style and texture guidance. Unlike CITIM, CCP performs global retrieval without explicit tumor masking, while the injected masks $\tilde{y}$ constrains the lesion layout during synthesis.

Conceptually, predicting target contrast features is a process of learning a mapping from available contrast to target contrast. This mapping is achieved using Adaptive Instance Normalization \cite{Alpher36}, which injects the statistics (mean $\mu$ and variance $\sigma$) of the retrieved prototype $F_{l^*}$ into the available feature $F^j_s$:
\begin{equation}
  \text{AdaIN}(F_{l^*}, F^j_s) = \sigma(F_{l^*}) \cdot \frac{F^j_s-\mu(F^j_s)}{\sqrt{\sigma^2(F^j_s)+\epsilon}} + \mu(F_{l^*})
  \label{eq:adain}
\end{equation}

Here, the retrieved prototype provides channel-wise statistics, while the spatial structure remains determined by the subject's own available features. This design improves target contrast texture without importing foreign anatomy.

This is followed by a Spatial-Channel Attention Block (CBAM) \cite{Alpher37} to refine structural consistency. This operation produces a set of candidate features $\{F_s^{'}\}$, which are fused via a $1\times 1$ convolution $\gamma$ to yield the predicted target feature:
\begin{equation}
  F_{\text{CCP}}^j = \gamma (\text{Concat}( F_1^{'}, \dots, F_s^{'}))
  \label{eq:fCPM}
\end{equation}

To align the representation space, an MSE loss is imposed between the predicted feature $F_{\text{CCP}}^j$ and the ground-truth feature $F_t^j$ (extracted by the encoder $E^{t}$ from the real target contrast $x^t$):
\begin{equation}
  \ell{_{f}} = \frac{1}{|T|} \sum_{t \in \mathcal{T}}  \big(\frac{1}{5} \sum_{j=1}^{5} ||F_{\text{CCP}}^j - F^j_t||^2 \big)
  \label{eq:lf}
\end{equation} 

To ensure the retrieval augmentation remains effective, we dynamically update two memory banks during training using a First-In-First-Out (FIFO) queue strategy. This prevents the "slow drift" phenomenon \cite{Alpher34} by discarding outdated features, ensuring the bank statistics remain aligned with the evolving encoder parameters.

\subsection{Loss function}

The segmentation branch is optimized using a combination of the CTDL loss, binary cross-entropy (BCE), and Dice loss. The total segmentation objective is defined as:
\begin{equation}
    \ell_{\text{seg}} = \alpha\ell_c + \ell_{\text{BCE}}(y, \tilde{y}) + \ell_{\text{Dice}}(y, \tilde{y})
  \label{eq:lseg}
\end{equation}

\begin{equation}
\ell_{\text{Dice}} = 1 - \frac{2 \cdot (y \cdot \tilde{y}) + \epsilon}{y + \tilde{y} + \epsilon}
\end{equation}

\begin{equation}
\ell_{\text{BCE}} = - \left(
    y \cdot \log\left(\tilde{y}\right) + (1 - y) \cdot \log\left(1 - \tilde{y}\right) 
\right)
\end{equation}

where $\alpha$ is a balancing hyper-parameter.

For the synthesis branch, the objective $\ell_{\text{syn}}$ comprises five terms: reconstruction loss ($L_1$), synthesis loss ($L_1$), feature loss (\Cref{eq:lf}), a tumor-focused local loss, and an adversarial loss (Least Squares GAN). The synthesis objective is formulated as:
\begin{equation}
  \begin{split}
    \ell_{\text{syn}} &= \beta_1\sum_{t \in \mathcal{T}} ||\tilde{x}^t - x^t||_1 + \beta_2\sum_{s \in \mathcal{S}} ||\tilde{x}^s - x^s||_1 \\
    &+ \beta_3\ell_f + \beta_4\sum_{i \in X} ||(\tilde{x}^i \odot y) - (x^i \odot y)||_1 \\
    &+ \beta_5 \sum_{i \in \mathcal{T}} (P_i(\tilde{x}^i) - 1)^2
  \end{split}
  \label{eq:lsyn}
\end{equation}
where $\beta_{1-5}$ are hyperparameters. The corresponding discriminator loss is defined as:
\begin{equation}
    \ell_{\text{dis}} = \sum_{i \in \mathcal{T}} \left[ (P_i(x^i) - 1)^2 + (P_i(\tilde{x}^i))^2 \right]
  \label{eq:ldis}
\end{equation}

\section{Experiments}
\subsection{Datasets}
To evaluate the effectiveness of our proposed framework, we conduct extensive experiments on two publicly available multi-contrast brain MRI datasets.

\textbf{Brain Tumor Segmentation Challenge 2020 (BraTs2020)}. The BraTs2020 dataset comprises 3D multi-contrast MRI scans (FLAIR, T1, T1ce, and T2) from 369 glioma patients. All scans are co-registered to the same anatomical template (SRI24), interpolated to a unified resolution of $1 \times 1 \times 1$ mm\textsuperscript{3}, and skull-stripped. The dataset provides pixel-level ground-truth tumor segmentation masks annotated and approved by expert neuroradiologists. Given that the ground truth masks for the official validation and test sets are withheld, we perform our experiments using the 369 training cases. Specifically, we randomly partition these cases into training (70\%), validation (10\%), and testing (20\%) sets to ensure a rigorous evaluation.

\textbf{UCSF-BMSR Dataset.} The UCSF-BMSR dataset is a large-scale clinical repository containing 560 multi-contrast MRI examinations from 412 patients undergoing Gamma Knife radiosurgery. This dataset is particularly challenging due to the small size and high variability of metastatic lesions. It includes expert voxel-wise annotations for 5,136 brain metastases across the T1-weighted contrast-enhanced sequences. Consistent with the BraTs2020 protocol, we employ a 70\%/10\%/20\% random split for training, validation, and testing.

\subsection{Implementation details}

\textbf{Training Details}: All volumes are normalized to [-1,1] and cropped from $240\times240$ to $192\times192$. For 2D training, we sample the 20 central slices with the largest tumor areas. The model is implemented in PyTorch and trained on one NVIDIA RTX 3090 GPU using mixed precision. We train for 600 epochs with batch size 2 using Adam optimizer, an initial learning rate of $2\times10^{-4}$, and betas of (0.5, 0.999). The learning rate linearly decays after 300 epochs.

The loss weights are set to $\alpha=0.5$, $\beta_1=100$, $\beta_2=30$, $\beta_3=50$, $\beta_4=50$, and $\beta_5=1$. The memory bank size is $L=300$, which balances feature diversity and computational efficiency.

\textbf{Evaluation Metrics}: Quantitative performance is assessed using three widely recognized metrics: Peak Signal-to-Noise Ratio (PSNR) \cite{Alpher30} and Structural Similarity Index (SSIM) \cite{Alpher31} for pixel-level fidelity, and Learned Perceptual Image Patch Similarity (LPIPS) \cite{Alpher32} for perceptual quality. Higher PSNR and SSIM values indicate better performance, while lower LPIPS values denote higher perceptual similarity to the ground truth.

\begin{table*}[t]
  \centering
  \caption{Quantitative results of the comparison study on the BraTS2020 dataset. The best PSNR, SSIM and LPIPS values are in bold.}
  \label{tab:bras20}
  \resizebox{\linewidth}{!}{
  \begin{tabular}{lcccccccccccc}
    \toprule
    & \multicolumn{12}{c}{BraTs2020} \\
    \cmidrule(lr){2-13}
    & \multicolumn{3}{c}{FLAIR} & \multicolumn{3}{c}{T1} & \multicolumn{3}{c}{T1ce} & \multicolumn{3}{c}{T2} \\
    \cmidrule(lr){2-4} \cmidrule(lr){5-7} \cmidrule(lr){8-10} \cmidrule(lr){11-13}
    Method & PSNR$\uparrow$ & SSIM$\uparrow$ & LPIPS$\downarrow$ & PSNR$\uparrow$ & SSIM$\uparrow$ & LPIPS$\downarrow$ & PSNR$\uparrow$ & SSIM$\uparrow$ & LPIPS$\downarrow$ & PSNR$\uparrow$ & SSIM$\uparrow$ & LPIPS$\downarrow$ \\
    \midrule
    TSF\cite{Alpher13}& 23.31 & 0.8079 & 0.1367 & 23.59 & 0.8543 & 0.1179 & 24.31 & 0.8429 & 0.1464 & 25.05 & 0.8640 & 0.1097 \\
    GAE\cite{Alpher11}& 23.12& 0.8013& 0.1594& 23.86& 0.8624& 0.1190& 24.61& 0.8501& 0.1437& 25.17& 0.8653& 0.1152\\
    MMGAN\cite{Alpher09}& 23.42& 0.8038& 0.1555& 23.30& 0.8598& 0.1196& 24.56& 0.8506& 0.1441& 24.77& 0.8624&0.1166\\
    UMMIS\cite{Alpher14}& 23.95& 0.8185& 0.1386& 23.89& 0.8604& 0.1169& 24.96& 0.8570& 0.1386& 25.17& 0.8664& 0.1031\\
    APT\cite{Alpher29}& 24.13& 0.8229& 0.1357& 24.79& 0.8618& 0.1113& 25.04& 0.8579& 0.1303& 25.32& 0.8688& 0.1024\\
    \bottomrule
    Ours& \textbf{24.21} & \textbf{0.8262} & \textbf{0.1333} & \textbf{25.04} & \textbf{0.8634} & \textbf{0.1096} & \textbf{25.21} & \textbf{0.8598} & \textbf{0.1228} & \textbf{25.55} & \textbf{0.8707} & \textbf{0.1005} \\
    \bottomrule
  \end{tabular}
  }
\end{table*}

\begin{table*}[t]
  \centering
  
  \caption{Quantitative results of the comparison study on tumor regions of the BraTs2020 dataset. The best PSNR, SSIM, and LPIPS values are highlighted in bold.}  
  \label{tab:bra20tumor}
  \resizebox{\linewidth}{!}{
  \begin{tabular}{lcccccccccccc}
    \toprule
    & \multicolumn{12}{c}{BraTs2020} \\  
    \cmidrule(lr){2-13}  
    & \multicolumn{3}{c}{FLAIR(tumor)} & \multicolumn{3}{c}{T1(tumor)} & \multicolumn{3}{c}{T1ce(tumor)} & \multicolumn{3}{c}{T2(tumor)} \\  
    \cmidrule(lr){2-4} \cmidrule(lr){5-7} \cmidrule(lr){8-10} \cmidrule(lr){11-13}  
    Method & PSNR$\uparrow$ & SSIM$\uparrow$ & LPIPS$\downarrow$ & PSNR$\uparrow$ & SSIM$\uparrow$ & LPIPS$\downarrow$ & PSNR$\uparrow$ & SSIM$\uparrow$ & LPIPS$\downarrow$ & PSNR$\uparrow$ & SSIM$\uparrow$ & LPIPS$\downarrow$ \\  
    \midrule
    TSF\cite{Alpher13}& 15.41 & 0.5148 & 0.0608 & 18.76 & 0.6364 & 0.0417 & 18.84 & 0.6078 & 0.0451 & 18.02 & 0.6032 & 0.0413 \\  
    GAE\cite{Alpher11}& 15.76& 0.4996& 0.0576& 19.15& 0.6529& 0.0436& 19.21& 0.6304& 0.0423& 18.19& 0.6210& 0.1152\\
 MMGAN\cite{Alpher09}& 12.06& 0.4523& 0.0691& 13.05& 0.5123& 0.0488& 12.95& 0.4060& 0.0648& 11.40& 0.4730&0.0473\\
 UMMIS\cite{Alpher14}& 15.66& 0.5223& 0.0584& 19.31& 0.6573& 0.0394& 19.20& 0.6242& 0.0407& 18.67& 0.6316& 0.0368\\
    APT\cite{Alpher29}& 15.72& 0.5042& 0.0596& 18.92& 0.6412& 0.0402& 19.07& 0.6157& 0.0442& 18.33& 0.6119& 0.0397\\
    \bottomrule
 Ours& \textbf{16.09}& \textbf{0.5336}& \textbf{0.0525}& \textbf{19.77}& \textbf{0.6616}& \textbf{0.0376}& \textbf{19.30}& \textbf{0.6329}& \textbf{0.0382}& \textbf{18.68}& \textbf{0.6440}& \textbf{0.0353}\\
 \bottomrule
  \end{tabular}
    }
\end{table*}

\begin{table*}[t]
  \centering

  \caption{Quantitative results of the comparison study on UCSF-BMSR dataset. The best PSNR, SSIM and LPIPS values are in bold.}  
  \label{tab:ucmr}
  \resizebox{\linewidth}{!}{
  \begin{tabular}{lcccccccccccc}
    \toprule
    & \multicolumn{12}{c}{UCSF-BMSR} \\  
    \cmidrule(lr){2-13}  
    & \multicolumn{3}{c}{FLAIR} & \multicolumn{3}{c}{T1} & \multicolumn{3}{c}{T1ce} & \multicolumn{3}{c}{T2} \\  
    \cmidrule(lr){2-4} \cmidrule(lr){5-7} \cmidrule(lr){8-10} \cmidrule(lr){11-13}  
    Method & PSNR$\uparrow$ & SSIM$\uparrow$ & LPIPS$\downarrow$ & PSNR$\uparrow$ & SSIM$\uparrow$ & LPIPS$\downarrow$ & PSNR$\uparrow$ & SSIM$\uparrow$ & LPIPS$\downarrow$ & PSNR$\uparrow$ & SSIM$\uparrow$ & LPIPS$\downarrow$ \\  
    \midrule
    TSF\cite{Alpher13} & 23.65& 0.8599& 0.1037& 23.64& 0.8761& 0.1024& 26.15& 0.8763& 0.1299& 25.32& 0.8606& 0.1287\\  
    GAE\cite{Alpher11}& 23.98& 0.8591& 0.1065& 24.29& 0.8715& 0.1248& 26.98& 0.8847& 0.1298& 25.90& 0.8862& 0.0997\\
 MMGAN\cite{Alpher09}& 24.14& 0.8730 & 0.1412 & 22.74 & 0.8649 & 0.1282 & 25.25 & 0.8767 & 0.1055 & 23.69 & 0.8609 & 0.1043 \\
 UMMIS\cite{Alpher14}& 24.26& 0.8763& 0.0933& 24.13& 0.8814& 0.0954& 26.37& 0.8822& 0.1179& 25.56& 0.8858&0.0864\\
    APT\cite{Alpher29}& 24.29& 0.8796& 0.1002& 24.22& 0.8822& 0.0938& 26.89& 0.8841& 0.1056& 25.84& 0.8863& 0.0823\\
    \bottomrule
 Ours& \textbf{24.37} & \textbf{0.8810} & \textbf{0.0925} & \textbf{24.30} & \textbf{0.8833} & \textbf{0.0928} & \textbf{27.05} & \textbf{0.8858} & \textbf{0.0979} & \textbf{25.97} & \textbf{0.8872} & \textbf{0.0798} \\
 \bottomrule
  \end{tabular}
  }
\end{table*}

\subsection{Comparison with State-of-the-art Methods}
We benchmark our method against five state-of-the-art approaches for multi-contrast brain MRI synthesis: TSF \cite{Alpher13}, which employs a task-specific fusion strategy for adaptive information integration; GAE \cite{Alpher11}, which utilizes a unified hyper-network for sequential synthesis and segmentation; MMGAN \cite{Alpher09}, a multi-modal GAN designed to impute missing sequences via a single generator; UMMIS \cite{Alpher14}, a unified framework offering a general solution to modality imputation; and APT \cite{Alpher29}, a recent diffusion-based model that ensures structural consistency via anatomy-prior-guided transformation.

\begin{figure*}
  \centering
  \includegraphics[width=0.8\linewidth]{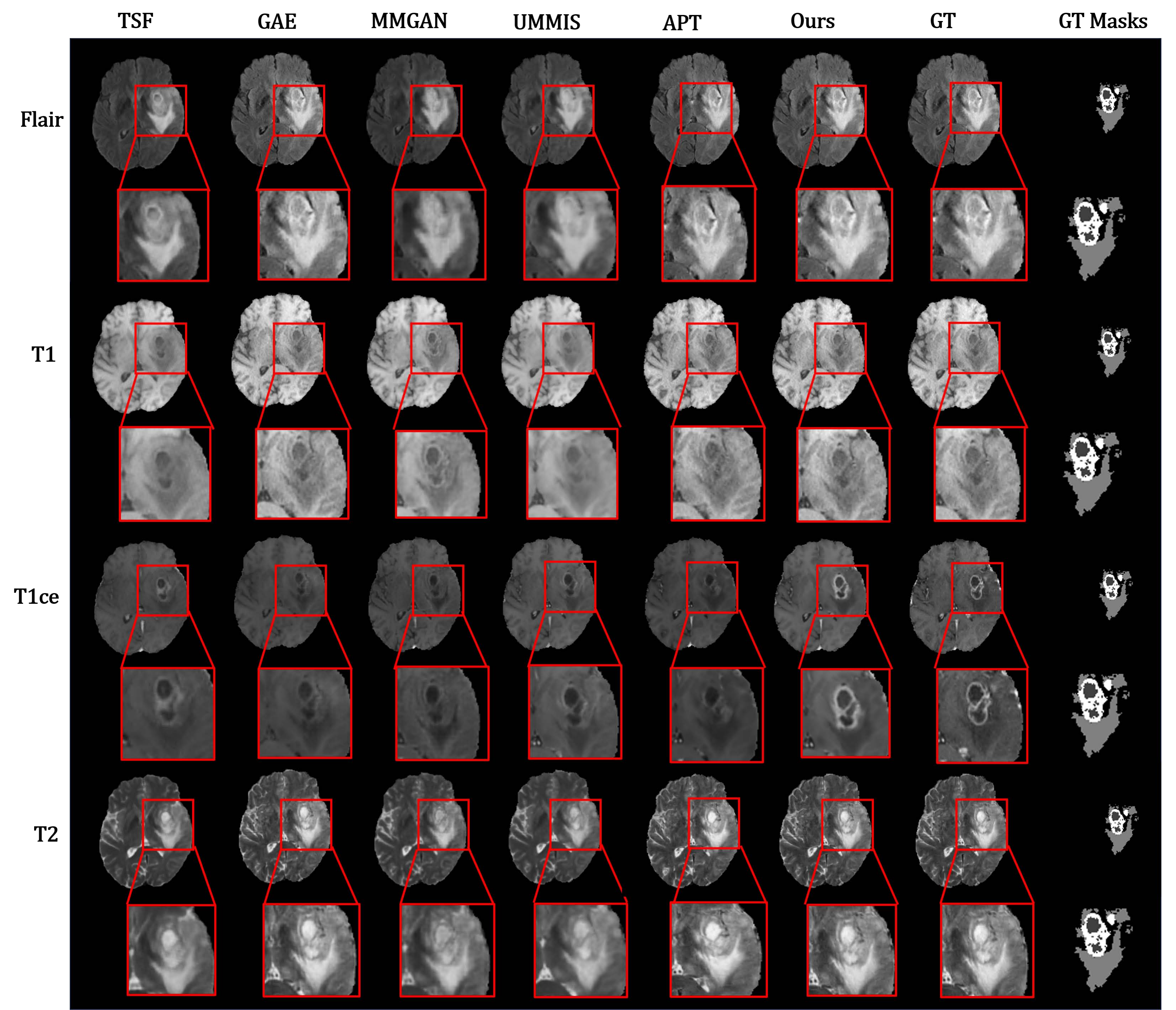}
  \hfill
  \caption{Qualitative results of comparison study on BraTS2020 dataset under the single contrast missing scenario. Red boxes indicate tumor regions.}
  \label{fig:bra2020}
\end{figure*}

Quantitative comparisons for BraTS2020 and UCSF-BMSR are presented in \Cref{tab:bras20} and \Cref{tab:ucmr} respectively. Our method consistently outperforms competing approaches across all metrics (PSNR, SSIM, and LPIPS). Qualitative results are illustrated in \Cref{fig:bra2020}. As shown, while competing methods often produce blurred textures or hallucinatory artifacts in tumor regions, our model preserves anatomical details more effectively, particularly in synthesizing complex tumor regions, demonstrating the efficacy of our dual-bank strategy and our method strictly follows the patient's anatomical layout, whereas retrieval does not alter the lesion boundaries, confirming that the dual-bank refines texture without corrupting structure. 

Although the quantitative improvements in global metrics (PSNR, SSIM) appear marginal compared to the second best method(APT), these values must be interpreted within the context of clinical utility. Global metrics are heavily biased towards the large volume of healthy brain tissue which is straightforward to synthesize, thereby obscuring improvements in small yet pathologically critical lesion regions. As evidenced in \Cref{tab:bra20tumor}, our method achieves a substantial performance boost over APT and other methods across all three metrics specifically within the tumor regions, highlighting its superior ability to preserve diagnostic detail.

\begin{table}
  \caption{Quantitative results of the ablation study on BraTs2020, averaged over all 14 missing contrast scenarios. }
  \label{tab:ablation1}
  \centering
  \begin{tabular}{@{}lc@{}lcc}
    \toprule
    Experiments& SSIM$\uparrow$    &&PSNR$\uparrow$ &LPIPS$\downarrow$\\
    \midrule
    Baseline & 0.8426 &&24.19&0.1376\\
 Baseline+seg(CITIM)& 0.8456 && 24.25&0.1282\\
    Baseline+CCP& 0.8472 && 24.48&0.1254\\
    \bottomrule
    Baseline+seg(CITIM)+CCP& \textbf{0.8548} && \textbf{25.01}&\textbf{0.1212}\\
    \bottomrule
  \end{tabular}
\end{table}

\subsection{Ablation study}
\textbf{Effectiveness of Synthesis Branch Components}. We begin with a baseline model comprising only the shared encoders and synthesis decoders. As shown in \Cref{tab:ablation1}, incorporating the auxiliary segmentation branch yields significant improvements in both PSNR and SSIM. This empirically validates our core hypothesis: just as contrast completion aids segmentation, the generation of accurate tumor masks can, conversely, provide critical semantic guidance to optimize synthesis quality. Furthermore, the integration of the Cross-Contrast Prediction (CCP) modules delivers additional performance gains. This improvement explicitly confirms the value of our cross-image multi-contrast feature memory bank. By retrieving and fusing relevant prototypes, the CCP module enables the network to generate target contrasts that are structurally complete and texturally realistic.

\textbf{Effectiveness of Segmentation Branch Design and Training Strategy}. We further investigate the design of the segmentation branch. We report the Dice for Whole Tumor (WT), Tumor Core (TC), and Enhancing Tumor (ET). Removing the Cross-Image Tumor Information Modeling (CITIM) module leads to a noticeable degradation in segmentation performance, underscoring the importance of leveraging cross-image tumor priors.

\begin{table}
  \caption{Dice results of the ablation study on Segmentation Branch.}
  \label{tab:ab1}
  \centering
  \small  
  \setlength{\tabcolsep}{4pt}  
  \begin{tabular}{lccc}  
    \toprule
    Experiments & WT & ET & TC \\  
    \midrule
    baseline & 0.8150 & 0.6194 & 0.7235 \\
    baseline+CITIM& 0.8168 & 0.6263 & 0.7340 \\
    baseline+CITIM+Syn(two stage)& 0.8037 & 0.6021 & 0.7142 \\
    baseline+CITIM+Syn(closed-loop)& \textbf{0.8173} & \textbf{0.6281} & \textbf{0.7423} \\
    \bottomrule
  \end{tabular}
\end{table}
\begin{table*}[t]
  \centering

  \caption{Impact of bank size on the BraTs2020 dataset. The best PSNR, SSIM and LPIPS values are in bold.}  
  \label{tab:bank}
    \resizebox{\linewidth}{!}{
  \begin{tabular}{lcccccccccccc}
    \toprule
    & \multicolumn{3}{c}{FLAIR} & \multicolumn{3}{c}{T1} & \multicolumn{3}{c}{T1ce} & \multicolumn{3}{c}{T2} \\  
    \midrule
    BankSize & PSNR$\uparrow$ & SSIM$\uparrow$ & LPIPS$\downarrow$ & PSNR$\uparrow$ & SSIM$\uparrow$ & LPIPS$\downarrow$ & PSNR$\uparrow$ & SSIM$\uparrow$ & LPIPS$\downarrow$ & PSNR$\uparrow$ & SSIM$\uparrow$ & LPIPS$\downarrow$ \\  
    \midrule
    L=10& 22.29& 0.8013& 0.1393& 23.44& 0.8547& 0.1142& 24.76& 0.8309& 0.1451& 24.10& 0.8549& 0.1192\\  
    L=100& 23.54& 0.8078& 0.1532& 23.96& 0.8605& 0.1173& 24.52& 0.8480& 0.1460& 24.71& 0.8570& 0.1152\\
 L=300& \textbf{24.21} & \textbf{0.8262} & \textbf{0.1333} & \textbf{25.04} & \textbf{0.8634} & \textbf{0.1096} & \textbf{25.21} & \textbf{0.8598} & \textbf{0.1228} & \textbf{25.55} & \textbf{0.8707} & \textbf{0.1005} \\
 L=500& 24.14& 0.8250& 0.1379& 24.89& 0.8623& 0.1119& 24.97& 0.8579& 0.1257& 25.39& 0.8694& 0.1021\\
    L=1000& 24.19& 0.8239& 0.1346& 24.88& 0.8627& 0.1103& 25.14& 0.8588& 0.1242& 25.43& 0.8688& 0.1013\\
    
  \bottomrule
  \end{tabular}
    }
\end{table*}

Moreover, we analyze the impact of our training strategy by comparing our end-to-end closed-loop scheme against a two-stage approach (where the segmentation network is pre-trained). As presented in  \Cref{tab:ab1}, the two-stage strategy results in inferior segmentation metrics. We attribute this to the fact that while a pre-trained segmentation branch maintains its original precision, joint optimization necessitates a trade-off: the segmentation branch's individual performance is slightly compromised as the shared encoders adapt to prioritize mutually beneficial representations that optimize the synthesis objective. In contrast, our closed-loop strategy not only boosts synthesis quality but also maintains and even slightly enhances segmentation accuracy, demonstrating the robustness and synergy of our unified framework.

\subsection{Parameter Sensitivity Analysis}
Memory Bank Size ($L$): As shown in \Cref{tab:bank}, we investigated the impact of bank size ranging from 10 to 1000. Performance peaks at $L=300$. A smaller bank ($L=10, 100$) lacks sufficient diversity to represent complex tumor patterns, while an excessively large bank ($L=500, 1000$) introduces redundant or noisy prototypes that may distract the retrieval process, confirming $L=300$ as the optimal balance between diversity and relevance.

Loss Weights ($\alpha, \beta$): regarding the loss weights in \Cref{eq:lseg} and \Cref{eq:lsyn}, we performed a grid search on a validation subset. We found that the segmentation weight $\alpha$ is sensitive; setting $\alpha < 0.1$ weakens the semantic guidance, leading to blurred tumor boundaries, while $\alpha > 1.0$ causes the encoder to over-specialize in segmentation, degrading synthesis texture. The synthesis weights $\beta_{1-5}$  were tuned to balance pixel-level consistency ($L_1$) and perceptual realism (adversarial loss). Specifically, a high weight on reconstruction ($\beta_1=100$) was necessary to stabilize the training of the shared encoder in the early stages.

\begin{table}
  \caption{Dice performance  on downstream tumor segmentation task. }
  \label{tab:example}
  \centering
  \begin{tabular}{@{}lc@{}lcc}
    \toprule
    Method & WT&&ET&TC\\
    \midrule
 Full contrasts& 0.8327& & 0.7631&0.8017\\
    TSF\cite{Alpher13}& 0.7302&&0.5468&0.6549\\
 GAE\cite{Alpher11}& 0.7672&& 0.5360&0.6382\\
    MMGAN\cite{Alpher09}& 0.7102&& 0.4968&0.6149\\
    UMMIS\cite{Alpher14}& 0.7921&& 0.5661&0.6661\\
 APT\cite{Alpher29}& 0.8021& & 0.5772&0.6823\\
    \bottomrule
         Ours& \textbf{0.8094}&& \textbf{0.5965}&\textbf{0.7105}\\
  \end{tabular}
\end{table}

\subsection{Downstream Task Evaluation}
 
To assess the clinical utility of the synthesized images, we employ tumor segmentation as a downstream validation task. Accurate segmentation of synthesized lesions is a rigorous proxy for diagnostic reliability. We utilize a standard U-Net, pre-trained on the complete set of ground-truth contrasts. For evaluation, we simulate all 14 possible missing contrast scenarios in the BraTs2020 dataset. For each scenario, the missing contrasts are synthesized by the competing methods and used to replace the ground truth as input to the pre-trained U-Net. The average segmentation results across all scenarios are summarized in \Cref{tab:example}. Our method achieves the highest segmentation accuracy, yielding substantial performance gains over state-of-the-art competitors. This indicates that our synthesized images preserve critical pathological semantics more faithfully, thereby offering greater reliability for downstream clinical analysis and computer-aided diagnosis.

\section{Conclusions}
In this work, we proposed a robust framework for any-to-any multi-contrast brain MRI synthesis, characterized by a unique segmentation-assisted closed-loop mechanism and a retrieval augmentation strategy. Our method effectively bridges the gap between global image generation and local lesion preservation. The incorporation of the segmentation branch provides explicit semantic guidance, while the dual-bank memory mechanism leverages cross-subject priors to reconstruct high-fidelity textures even when multiple contrasts are missing. Comprehensive evaluations on two public datasets validate the superiority of our approach over state-of-the-art methods, particularly in recovering complex tumor structures. These results suggest that our framework holds significant potential for enhancing downstream clinical tasks, such as tumor quantification and treatment planning, where complete multi-contrast data is often unavailable. Future work will explore extending this retrieval-based paradigm to other anatomical regions and 3D volumetric synthesis tasks with higher efficiency.

%
%
%
\bibliographystyle{splncs04}
\bibliography{mybibliography}
%




\end{document}